\documentclass[10pt,twocolumn,letterpaper]{article}
\usepackage{newtxtext}
\usepackage{cvpr}
\usepackage{multirow}
\usepackage{iftex}
\usepackage[hypcap=false]{caption}
\ifPDFTeX\else
  \usepackage{fontspec}
\fi
\usepackage{tikz}
\usetikzlibrary{positioning,arrows.meta}
\definecolor{cvprblue}{rgb}{0.21,0.49,0.74}
\usepackage[pagebackref,breaklinks,colorlinks,allcolors=cvprblue]{hyperref}

\def\paperID{*****}
\def\confName{CVPR}
\def\confYear{2027}

\newcommand{\ours}{\textsc{PassCal}}
\newcommand{\pass}{\textsc{passable}}
\newcommand{\blocked}{\textsc{blocked}}
\newcommand{\depends}{\textsc{depends}}

\newcommand{\AuditorKEightGain}{0.034}
\newcommand{\AuditorKEightCILow}{0.018}
\newcommand{\AuditorKEightCIHigh}{0.052}
\newcommand{\AuditorKEightP}{0.026}
\newcommand{\AuditorKTwentyGain}{0.043}

\title{Passable for Whom? Few-Shot Learning for Personal Calibration of Sidewalk Accessibility Perception for Human-Centered Autonomous Driving}

\author{Wesley Maia\quad Angel Martinez-Sanchez\quad Maitrayee Keskar\quad Ross Greer\\University of California, Merced}

\makeatletter
\apptocmd{\@maketitle}{%
  \vspace{10pt}
  \begin{center}
    \includegraphics[width=0.95\linewidth]{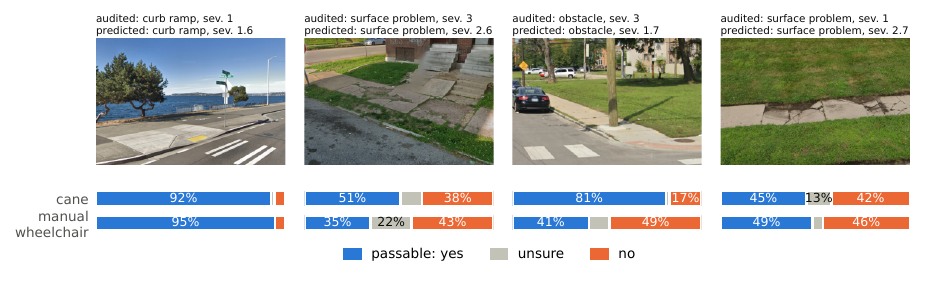}
  \end{center}
  \captionof{figure}{\textbf{The same sidewalk, different answers.} Four survey scenes with the answers of cane users and manual-wheelchair users, and the properties recorded by Project Sidewalk auditors (\emph{audited}) and by our detector (\emph{predicted}). People disagree between aids and within an aid. \ours{} describes the scene and learns from a few answers what that description means for one person.}
  \label{fig:teaser}
  \vspace{1.2em}
}{}{}
\makeatother

\begin{document}
\maketitle

\begin{abstract}
Whether a sidewalk is passable depends on who is trying to pass as people who use the same mobility aid disagree about the same sidewalk image. A single label per scene or mobility-aid group therefore
misrepresents many individuals' judgments. We introduce few-shot personal accessibility perception where given $K$ answers from a new person, it predicts their judgment of sidewalk scenes never seen before. \ours{} separates a shared description of the scene from the person's interpretation of it. The description comes from the internal state of a frozen vision-language model (VLM) and from barrier type, severity and context properties predicted by a detector trained on $14.9$k public audits; the interpretation, a calibration of the person's response style and a weighting of the properties, is fit from their answers. On $8{,}883$ judgments in which each of $135$ verified mobility-aid users rated the same $50$ sidewalk scenes, with people and scenes both held out, eight answers lower the Brier score from $0.550$ to $0.479$. The same VLM prompted with those answers reaches $0.562$, and the same calibration without the scene $0.521$. Conformal risk control turns predictions into \emph{passable}, \emph{blocked} or \emph{depends on you} statements with a bounded expected rate of wrong statements over people; at $5\%$, eight answers raise the share of decisive scenes from $27\%$ to $42\%$. Audit-supervised properties improve all sixteen vision and VLM encoders we test and outperform attributes from a sidewalk navigation VLM.
\end{abstract}

\section{Introduction}
\label{sec:intro}

A missing curb ramp stops one wheelchair user from egress and is a short detour for another. A cracked slab that a cane user steps over is a hazard for someone using a walker. Accessibility maps and routing tools~\cite{bolten2017accessmap,saha2022mypath,tannert2018wheelchair,openrouteservice2026wheelchair} nevertheless attach one label to each place, and learned perception systems~\cite{omeara2025rampnet,10.1145/3663548.3688531,hosseini2023tile2net} are trained to reproduce that label. Li et al.~\cite{DBLP:journals/corr/abs-2502-19888} showed images of sidewalk barriers to people who use canes, walkers, mobility scooters, and manual and motorized wheelchairs, and found large differences between these groups. In their data we find that differences \emph{within} a group are as large (\cref{fig:teaser}): for cane and walker users, a person's own tendency explains more of the variance in their answers than the image does ($27\%$ against $14\%$ and $18\%$; supplement).

Judgments like these are hard to collect. In the case of \cite{DBLP:journals/corr/abs-2502-19888}, each one comes from a person who actually uses a cane, walker, wheelchair or scooter, reached through disability organizations over two months, paid for a $25$-minute survey, and screened against the automated and incentive-driven sign-ups that paid disability studies attract: Li et al. removed $5{,}239$ fraudulent sign-ups and $68$ suspicious completed surveys to obtain theirs~\cite{DBLP:journals/corr/abs-2502-19888}. 

The resulting $8{,}883$ answers represent about $56$ hours of mobility-aid users' time, from more people than in most accessibility studies ( median participants at CHI and ASSETS conferences is $13$~\cite{mack2021accessibility}). Each person rates the same $50$ scenes, allowing differences between people to be distinguished from differences between scenes. It is, to our knowledge, the only public dataset of individual sidewalk passability judgments by mobility-aid users. Every new scene must be judged by such people, so the cost lies in the answers, not in the images. Methods should therefore be judged by how accurately they predict a person's judgments from only a few of their answers.

This paper asks how can a vision system predict an individual's accessibility judgments when population labels overlook personal differences and collecting personal judgments is costly. We separate a shared description of the scene from a person's interpretation of it. Public audits and a frozen vision-language model (VLM) supply the description; a short interaction supplies the personal answers that an aid label cannot determine. The target is the judgment a person reports about a photograph. The formulation has three parts.
\begin{enumerate}[leftmargin=*,itemsep=1pt,topsep=2pt]
    \item \textbf{Perception.} A detector trained on public Project Sidewalk audits~\cite{Saha_ProjectSidewalk_CHI2019} predicts the barrier type, its severity, and context tags for a street-view image, and a frozen VLM asked whether the sidewalk is passable provides a scene embedding. No user judgment is involved.
    \item \textbf{Personal calibration.} A person answers $K$ questions about calibration images only once, with no further answers needed when predicting their judgments on new scenes. From these answers we estimate personal weights on the properties and a calibration of the person's response style.
    \item \textbf{Statements with controlled risk.} The system outputs ``passable'', ``blocked'', or ``depends on you'', while retaining the underlying predicted probabilities for downstream use. Conformal risk control~\cite{angelopoulos2024crc} bounds how often each kind of statement is wrong, in expectation over people.
\end{enumerate}

We evaluate with both the person and the query images excluded from model fitting, which tests whether answers on calibration images improve predictions on other scenes, avoiding overfitting to a familiar rater or a scene's crowd label. Our method, \ours{}, learns how a person departs from a model of everyone else, with three response parameters and ten property weights. The two obvious baseline alternatives do not: despite being given eight user answers in its prompt, a recalibrated VLM is no better than a model that ignores the person, although the same frozen VLM used as a scene encoder inside \ours{} gives our best result; and a regressor fit per person on the VLM state, as done for personal aesthetics~\cite{ryu2026piaa}, cannot learn a person from eight answers.

Our contributions are:
\begin{itemize}[leftmargin=*,itemsep=1pt,topsep=2pt]
    \item A benchmark for few-shot personal accessibility perception: $8{,}883$ dense judgments from $135$ mobility-aid users, a protocol that holds out people and scenes, intervals that resample both, a held-out barrier type, and baselines from in-context VLMs, personal aesthetics, crowd oracles and a pedestrian navigation VLM. We will release code and splits.
    \item \ours{}, which separates an audit-supervised scene description from a few-shot personal interpretation. Eight answers reduce the Brier score by $0.070$, and \ours{} outperforms in-context personalisation of the same VLM by $0.090$ and per-person regression on its state by $0.068$.
    \item Evidence that audit supervision adds what foundation models miss: predicted sidewalk properties improve all sixteen vision and VLM encoders we test, and an oracle study locates the remaining gains in barrier type and severity.
    \item Statements whose expected rate of wrong answers per scene is bounded over people; at a $5\%$ bound, eight answers raise the decisive share (scenes labeled passable or blocked) from $27\%$ to $42\%$.
\end{itemize}

\section{Related Work}
\label{sec:related}

\textbf{Sidewalk accessibility from street-level imagery.} Crowdsourced audits of Street View~\cite{10.1145/2470654.2470744,10.1145/2642918.2647403,Saha_ProjectSidewalk_CHI2019} have been scaled with classifiers and detectors~\cite{10.1145/3308561.3353798,10.1145/3663548.3688531,omeara2025rampnet} and complemented by aerial extraction~\cite{hosseini2023tile2net,hosseini2022citysurfaces} and multimodal sensing~\cite{damaceno2024sideseeing}. Recent studies test vision-language models on accessibility attributes: they estimate sidewalk width reasonably but slope and pavement poorly~\cite{lieu2026vlmsidewalk}, and miss subtle surface conditions~\cite{wang2026vlmsensors}, and grounded VLMs now converse about pedestrian navigation~\cite{sultan2026walkgpt}. These works predict a property of the place, but we predict the judgment of a person.

\textbf{Personalised accessibility.} Routing systems accept hand-set user weights~\cite{bolten2017accessmap,saha2022mypath,tannert2018wheelchair}. Li et al.~\cite{DBLP:journals/corr/abs-2502-19888} collected the judgments we use and analysed differences between mobility groups statistically, without a predictive model of individuals. Accessibility Scout~\cite{huang2025accessibilityscout} personalises LLM-generated scans of photos and evaluates them on $500$ images and in a $10$-person user study, without a predictive model of individual judgments. We contribute a quantitative protocol and a model at the individual level.

\textbf{Personalised and subjective visual prediction.} Personalised image aesthetics adapts a generic model with a few ratings per user~\cite{ren2017personalized,yang2022para,yun2024taskvector,yang2026prac}; Ryu and Yanaka~\cite{ryu2026piaa} fit per-user linear heads on the hidden states of a frozen VLM, the closest analogue to our use of a VLM state. Few-shot personalisation also predicts an unseen person's gaze~\cite{xue2025scanpath,xue2025deper}, and a few answers from a new user reweight shared attributes or earlier users' preferences in generation and interface design~\cite{bai2026testtime,peng2026genui}. We differ in three ways: the shared description is supervised by audits of the property in question, the person's response style is calibrated separately from their weights, and the output is a statement with a bounded error rate, evaluated on people and images that are both new. Individual quality models learn observer-specific responses to distortions~\cite{fotiotiotsop2024individual}, and individual preferences are represented unevenly~\cite{goree2023correct}. Urban perception has moved from population scores~\cite{dubey2016deep} to differences across demographic groups~\cite{quintana2025specs} and to models conditioned on rater attributes~\cite{damandeh2026walkable}. We condition on a person's answers instead of their attributes, in a domain where an error has a physical cost. Work on learning from disagreement treats annotator variation as signal~\cite{peterson2019humanuncertainty,collins2022elicit,davani2022disagreements}; ours is the few-shot, unseen-annotator case.

\textbf{Concept-based models.} Concept bottleneck models predict through human-readable concepts~\cite{koh2020cbm}, with a residual on the embedding~\cite{yuksekgonul2023pcbm} or interactive correction~\cite{chauhan2023interactive,steinmann2024intervene}. Our concepts come from an external audit corpus, and what is personal is the weight a user gives them.

\textbf{Calibration and conformal prediction.} Temperature scaling corrects the confidence of a trained classifier~\cite{guo2017calibration}; we fit such a correction per person from a few answers. Conformal risk control bounds the expectation of a monotone loss with exchangeable calibration data~\cite{angelopoulos2024crc}, few-shot variants study per-task guarantees~\cite{park2023metaxb}, conformal intervals have been attached to VLM estimates of sidewalk attributes~\cite{lieu2026vlmsidewalk}, and group-conditional versions control risk within each group of a hierarchy with abstention~\cite{salem2026hgcrc}. We apply it with people as the exchangeable units.

\section{Data}
\label{sec:data}

\textbf{Judgments.} We use the public release of Li et al.~\cite{DBLP:journals/corr/abs-2502-19888}. Respondents saw street-view images of sidewalk barriers taken from Project Sidewalk and answered, for the mobility aid they use, whether they could pass: yes, unsure, or no. After removing $5{,}239$ fraudulent sign-ups and $68$ suspicious completed surveys, the release keeps $190$ survey responses. These include $178$ person-aid profiles from $135$ people: walking cane $53$, manual wheelchair $37$, motorized wheelchair $34$, walker $32$, mobility scooter $22$.Multi-aid users have separate profiles but remain one person for splitting and evaluation. We use the $50$ images that we could match to their Project Sidewalk record, with $8{,}883$ judgments ($50\%$ yes, $38\%$ no, $12\%$ unsure).

\textbf{Objective properties.} Each survey image is a Project Sidewalk label, which carries a label type (curb ramp, missing curb ramp, obstacle, surface problem), a severity from $1$ to $3$, and free tags. We group tags into five (narrow, vegetation, surface damage, object, no alternate route), which gives a $10$-dimensional property vector. We call the values recorded by auditors \emph{audited} properties and the detector's outputs \emph{predicted} properties.

\textbf{How people differ.} In only $36$ of the $260$ image-aid cells do at least $80\%$ agree. In \cref{fig:teaser}, cane users largely agree on the obstacle, while manual-wheelchair users are split. Higher severity lowers passability for every aid, yet for surface problems, narrow sidewalks and objects, $30$ to $60\%$ of people have a personal effect opposite to that of their group.

\textbf{Audit pool.} To train the property detector we sampled Project Sidewalk labels from nine cities that include the eight cities of the survey images through its public API, stratified by city, type and severity. We excluded every survey label and every label on a panorama that contains one, and kept labels whose panorama still resolves in the Street View Static API. The pool has $14{,}893$ scenes; an extension to $18{,}092$ is used in the scaling analysis and in \cref{sec:auditors}.

\section{Method}
\label{sec:method}

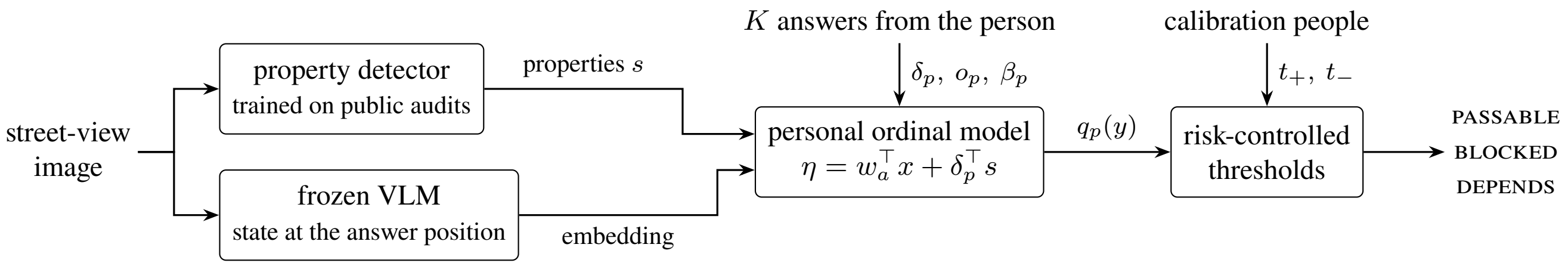
\begin{figure*}[t]
\centering
\begin{tikzpicture}[font=\small, node distance=5mm and 9mm,
  box/.style={draw, rounded corners=2pt, align=center, inner sep=4pt, minimum height=10mm},
  io/.style={align=center, inner sep=2pt},
  arr/.style={-{Stealth[length=5pt]}, semithick}]
  \node[io] (img) {street-view\\image};
  \node[box, right=of img, yshift=7mm] (det) {property detector\\{\footnotesize trained on public audits}};
  \node[box, right=of img, yshift=-7mm] (enc) {frozen VLM\\{\footnotesize state at the answer position}};
  \node[box, right=30mm of det, yshift=-7mm] (ord) {personal ordinal model\\$\eta = w_a^{\top}x + \delta_p^{\top}s$};
  \node[box, right=14mm of ord] (crc) {risk-controlled\\thresholds};
  \node[io, right=of crc] (out) {\textsc{passable}\\\textsc{blocked}\\\textsc{depends}};
  \node[io, above=7mm of ord] (ans) {$K$ answers from the person};
  \node[io, above=7mm of crc] (cal) {calibration people};
  \draw[arr] (img.east) -- ++(4mm,0) |- (det.west);
  \draw[arr] (img.east) -- ++(4mm,0) |- (enc.west);
  \draw[arr] (det.east) -- node[above] {\footnotesize properties $s$} ++(22mm,0) |- ([yshift=2mm]ord.west);
  \draw[arr] (enc.east) -- node[below] {\footnotesize embedding} ++(22mm,0) |- ([yshift=-2mm]ord.west);
  \draw[arr] (ans) -- node[right] {\footnotesize $\delta_p,\ o_p,\ \beta_p$} (ord);
  \draw[arr] (cal) -- node[right] {\footnotesize $t_{+},\ t_{-}$} (crc);
  \draw[arr] (ord) -- node[above] {\footnotesize $q_p(y)$} (crc);
  \draw[arr] (crc) -- (out);
\end{tikzpicture}
\caption{\textbf{\ours{}.} The detector and the frozen VLM describe the scene without any user judgment. A few answers from the person set the personal parameters, and thresholds calibrated on other people turn the prediction into a statement with bounded risk.}
\label{fig:method}
\end{figure*}

\subsection{Property detector}
A Project Sidewalk label is a point in a wide scene. From the label's position on the panorama we compute its pixel in the downloaded $90^\circ$ view (median error $0.1$\,px against the auditor's recorded click, on $6{,}000$ labels) and take a $50^\circ$ crop centred on it. A second crop, from the half of the image away from the label, serves as a background example. We fine-tune the last four blocks of DINOv2-L~\cite{oquab2023dinov2} with three heads: five-way type (four label types plus background), three-way severity on label crops, and five binary tags. Validation is grouped by panorama.

At test time the label location is not used, since a deployed system would not have it. The image is covered by square windows at three scales. Each window is scored, and per-image properties are the mean of the three highest-scoring windows per output, with severity weighted by the probability that a window is not background. We average three seeds.

\subsection{Scene embedding}
\label{sec:embedding}
The properties name what auditors record, not everything that bears on passability. For the rest we use a frozen VLM, Qwen3-VL-8B~\cite{bai2025qwen3vl}, as an encoder. The model receives the image and a question asking whether a person who uses a mobility aid could pass through this sidewalk (full prompt in the supplement), and we keep the last-layer hidden state at the position where it would begin its answer, discarding the answer itself. A $16$-dimensional PCA of this state, fit on $3{,}000$ pool scenes, is the scene embedding $z_i$. The VLM never sees a user's answers and is not fine-tuned. \Cref{sec:encoders} compares this choice with fifteen other encoders.

\subsection{Personal model}
\label{sec:model}
Let $x_i=(z_i,s_i)\in\mathbb{R}^{26}$ represent image $i$: the scene embedding concatenated with the $10$ predicted properties $s_i$. For a person $p$ using aid $a$, the answer follows a proportional-odds model~\cite{mccullagh1980ordinal} with latent passability
\begin{equation}
\eta_{p,i} = w_a^{\top} x_i + \delta_p^{\top} s_i
\label{eq:eta}
\end{equation}
and aid-specific thresholds $\tau_{a,0}<\tau_{a,1}$, so that $P_{p,i}(y\le c)=\sigma(\tau_{a,c}-\eta_{p,i})$ for $c\in\{\text{no},\text{unsure}\}$. The group weights $w_a$ and the thresholds are fit on training people. The group term uses the whole representation; the personal weights $\delta_p$ act only on the properties, so that what the system learns about how a person sees a scene is expressed over named quantities.

People also differ in how they respond, independently of the scene: some say no more readily, some rarely say unsure, some are more consistent than the group. We model this with a personal response calibration of the predicted distribution,
\begin{equation}
q_{p,i}(c) \propto \exp\!\big(\beta_p \log P_{p,i}(c) + o_{p,c}\big),
\label{eq:response}
\end{equation}
with class offsets $o_p$ that sum to zero and a positive inverse temperature $\beta_p$, a per-person form of temperature scaling~\cite{guo2017calibration}. The offsets capture strictness and the use of ``unsure''; $\beta_p$ captures confidence.

Both sets of personal parameters are estimated by MAP from the person's $K$ calibration answers, first the property weights,
\begin{equation}
\min_{\delta_p}\; -\!\sum_{k=1}^{K}\log P_{p,k}(y_{p,k}) + \lambda_\delta \lVert\delta_p\rVert^2 ,
\label{eq:map}
\end{equation}
and then $(o_p,\beta_p)$ on the same answers with Gaussian priors centred on $o_p{=}0$ and $\beta_p{=}1$. With $K{=}0$ the model returns the group prediction. The prior strengths are tuned on training people only. \Cref{sec:adapters} compares this choice with simpler personal parameterisations.

\subsection{Statements with controlled risk}
\label{sec:crc}
The system says \pass{} if $q(\text{yes})\ge t_{+}$, \blocked{} if $q(\text{no})\ge t_{-}$, and \depends{} otherwise. The harmful error for \pass{} is saying it to a person who answers no. For a person $p$ with held-out scenes $S_p$, define
\begin{equation}
L_p(t)=\frac{1}{|S_p|}\sum_{i\in S_p}\mathbf{1}\{q_{p,i}(\text{yes})\ge t\}\,\mathbf{1}\{y_{p,i}=\text{no}\},
\end{equation}
which is non-increasing in $t$. Given $n$ calibration people who went through the same $K$-answer procedure, conformal risk control~\cite{angelopoulos2024crc} selects the smallest $t_{+}$ with $\tfrac{n}{n+1}\bar{L}(t_{+})+\tfrac{1}{n+1}\le\alpha$, which guarantees $\mathbb{E}[L_{\text{new}}(t_{+})]\le\alpha$ for a new exchangeable person. The threshold $t_{-}$ is set symmetrically for saying \blocked{} to a person who answers yes; if both conditions hold, \pass{} takes precedence, which can only remove \blocked{} statements, so its bound still holds. An answer of unsure counts as neither error. The guarantee is marginal over people, not conditional on a person or an aid. Each statement type is bounded separately at $\alpha$, and the denominator is all of a person's scenes, not only the scenes that receive that statement. And calibration people are scored on the same held-out images as test people, so the bound covers new people on a fixed set of scenes. We report the \emph{decisive share}, the fraction of scenes that receive \pass{} or \blocked{}, at fixed $\alpha$.

\section{Experiments}
\label{sec:experiments}

\subsection{Protocol}
\label{sec:protocol}
People and images are each split into five folds. In fold $f$ the group model is fit on training people and training images. A test person gives $K$ answers on training images, as with a fixed calibration set shipped with an application, and is evaluated on the held-out images, whose votes were excluded from group-model fitting. The metric is the Brier score of the predicted distribution against the person's answer (summed over the three classes), averaged within a person and then over people. We repeat the whole procedure for $20$ random partitions of people and images, with three draws of calibration items each, and report means over partitions; the standard deviation of a table entry across partitions is $0.007$ to $0.010$. We compare methods using each person's mean score across partitions, with paired bootstrap $95\%$ confidence intervals over the $135$ people and Wilcoxon signed-rank tests. Because all partitions share the same $50$ images, we also report a two-way bootstrap that resamples people and images together (supplement), and a split that withholds a whole barrier type (\cref{sec:perception}). For statements, the non-test people are split in half into people who fit the model and people who calibrate thresholds, repeated $20$ times. Unless stated otherwise, every number in this section comes from this protocol.

\textbf{References.} Three variants modify \ours{}. \emph{Embedding only} removes predicted properties from the group model and moves personal weights to the embedding dimensions. \emph{Audited properties} replaces predicted properties with auditors' values. \emph{WalkGPT attributes} replaces them with ten attributes extracted from WalkGPT-13B's answers~\cite{sultan2026walkgpt} (supplement). Three baselines provide alternative personalisation approaches. \emph{No image} applies the response calibration of \cref{eq:response} to the aid's overall answer distribution, ignoring the image. \emph{Per-person regression} follows Linear-Hidden~\cite{ryu2026piaa}: a ridge regressor on the frozen VLM state, fit only on the person's answers. \emph{In-context VLM} shows Qwen3-VL-8B the person's eight calibration images and answers, then queries the test image; we read the next-token distribution over the three answers. Two oracles use votes on the test image unavailable at deployment. \emph{Crowd oracle} predicts the answer distribution of the $18$ to $42$ training people with the same aid. \emph{Crowd oracle + strictness} adds personal strictness to that distribution.

\subsection{Main results}

\begin{table}[t]
\centering
\small
\setlength{\tabcolsep}{4.5pt}
\begin{tabular}{lcccc}
\toprule
 & \multicolumn{4}{c}{Brier on the person's answers $\downarrow$} \\
Model & $K{=}0$ & $K{=}3$ & $K{=}8$ & $K{=}20$ \\
\midrule
Per-person regression~\cite{ryu2026piaa} & -- & 0.566 & 0.550 & 0.521 \\
No image + response cal. & 0.583 & 0.548 & 0.521 & 0.503 \\
WalkGPT~\cite{sultan2026walkgpt} attributes & 0.556 & 0.519 & 0.490 & 0.473 \\
Embedding only & 0.552 & 0.514 & 0.483 & 0.464 \\
\ours{} (ours) & 0.550 & 0.512 & \textbf{0.479} & \textbf{0.457} \\
\midrule
Audited properties & 0.537 & 0.497 & 0.463 & 0.438 \\
Crowd oracle & \multicolumn{4}{c}{0.514} \\
Crowd oracle + strictness & 0.514 & 0.474 & 0.450 & 0.437 \\
\bottomrule
\end{tabular}

\caption{New people on new sidewalks: $135$ people, $50$ images, mean over $20$ partitions. Per-person regression uses the first calibration draw (\ours{} scores $0.514$, $0.480$ and $0.458$ on the same items). Rows below the line use information that a deployed system does not have.}
\label{tab:main}
\end{table}

\begin{figure}[t]
\centering
\includegraphics[width=\linewidth]{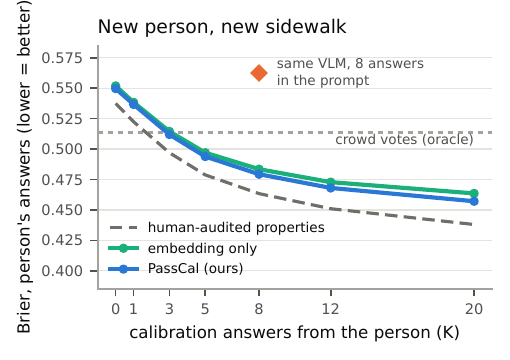}
\caption{Brier score against the number of calibration answers (mean over $20$ partitions). The diamond is the same VLM with the eight answers in its prompt, after recalibration (five partitions).}
\label{fig:curve}
\end{figure}

\textbf{Personalisation works on unseen sidewalks.} \Cref{tab:main} and \cref{fig:curve} give the main result. Eight answers reduce the Brier score of \ours{} by $0.070$ (CI $[0.051, 0.092]$, better for $68\%$ of people) and twenty by $0.093$ ($79\%$), in all $20$ partitions; resampling images as well as people leaves both intervals clear of zero (supplement). The mean score is below the crowd oracle's by $0.034$ at $K{=}8$ and $0.057$ at $K{=}20$; the interval excludes zero at both, also when images are resampled. Adding a personal strictness to the oracle, which then uses both the person's answers and everyone's votes on the test scene, gives the reference row of \cref{tab:main}. \Cref{sec:who} shows who gains most.

\textbf{Predicted properties help.} \ours{} is better than the embedding alone by $0.004$ at $K{=}8$ (better for $73\%$ of people) and $0.006$ at $K{=}20$ ($74\%$), in all $20$ partitions. With images resampled the difference stays clear of zero at $K{=}8$ ($[0.0003, 0.008]$) and just reaches it at $K{=}20$. Attributes parsed from WalkGPT~\cite{sultan2026walkgpt}, a VLM for pedestrian navigation, make the embedding worse instead ($0.490$ at $K{=}8$; \ours{} better by $0.011$, two-way CI $[0.003, 0.018]$): it names curbs, vehicles and vegetation in all fifty scenes and never a curb ramp or surface problem. Audited properties are ahead by $0.016$ and $0.019$; \cref{sec:perception} locates this gap.

\textbf{A per-person model does not suffice.} The Linear-Hidden model of Ryu and Yanaka~\cite{ryu2026piaa}, fit only on that person's eight answers, reaches a Brier score of $0.550$ on the $4096$-dimensional VLM state and $0.539$ on our $26$-dimensional representation, against $0.480$ for \ours{} on the same items (better for $93\%$ of people), and its per-person rank correlation with the answers is $0.10$ against $0.31$. With eight answers a person cannot be learned from scratch; what can be learned is how the person departs from a model of everyone else. The personal adapters of \cref{sec:adapters} are such shrinkage regressions, a hierarchical model with the group model as prior.

\textbf{Prompting the same VLM does not personalise.} On five partitions with identical people, calibration answers and test images, the VLM with eight in-context answers has a raw Brier score of $0.766$: its answer probabilities are poorly calibrated. We therefore recalibrate it with a per-aid multinomial logistic map fit on eight-answer prompts built from training people and training images of the same fold, so that the map matches the test condition and uses no test person or test image. This gives $0.562$, against $0.472$ for \ours{} on the same partitions ($-0.090$, CI $[-0.105,-0.076]$, $p{=}3{\times}10^{-19}$, better for $84\%$ of people). The recalibrated VLM with eight answers is worse than our model with none ($0.548$, $p{=}0.02$); the $32$B model gives $0.563$. Neither more examples nor a summary of the person's answers changes this: with twenty answers in the prompt the recalibrated VLM scores $0.546$ against $0.454$ for \ours{} ($p{=}5{\times}10^{-21}$, better for $89\%$ of people), and stating how often the person answered yes, unsure and no gives $0.556$ against $0.472$ (supplement). On these judgments the VLM does not extract the personal signal from the examples, although its hidden state describes the scene well enough for \ours{} to do so; Ryu and Yanaka~\cite{ryu2026piaa} likewise find a ten-shot prompt far below per-user heads for personal aesthetics.

\subsection{Statements with controlled risk}

\begin{table}[t]
\centering
\small
\setlength{\tabcolsep}{4pt}
\begin{tabular}{lcccccc}
\toprule
 & \multicolumn{3}{c}{$\alpha=0.05$} & \multicolumn{3}{c}{$\alpha=0.10$} \\
Model & $K{=}0$ & $8$ & $20$ & $K{=}0$ & $8$ & $20$ \\
\midrule
Embedding only & 27.2 & 41.0 & 45.8 & 57.0 & 75.3 & 78.5 \\
\ours{} & 26.6 & 41.9 & 46.2 & 57.6 & 76.0 & 79.4 \\
\midrule
Audited properties & 29.7 & 44.1 & 49.0 & 61.0 & 79.8 & 84.1 \\
Crowd oracle & \multicolumn{3}{c}{31.5} & \multicolumn{3}{c}{66.2} \\
\bottomrule
\end{tabular}

\caption{Decisive share (\%): scenes that receive \pass{} or \blocked{} when the expected frequency of each wrong statement is bounded at $\alpha$ ($20$ repetitions).}
\label{tab:decisive}
\end{table}

\begin{figure}[t]
\centering
\includegraphics[width=\linewidth]{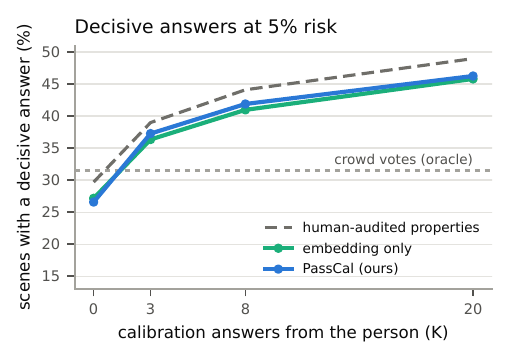}
\caption{Decisive share at a $5\%$ bound per statement type, against the number of calibration answers.}
\label{fig:guarantee}
\end{figure}

\Cref{tab:decisive} and \cref{fig:guarantee} report the decisive share. The realised frequency of a wrong \pass{} per scene is $0.035$ at $\alpha{=}0.05$ and $0.085$ at $\alpha{=}0.10$, within the bound; among the scenes that receive \pass{} at $\alpha{=}0.05$, $13.6\%$ are answered no, a ratio the bound does not control (\cref{sec:limits}). Without calibration answers the system commits on $27\%$ of scenes, and with eight on $42\%$ ($+15.3$ points paired, $p{=}6{\times}10^{-13}$, more decisive for $74\%$ of people). The comparison with the crowd oracle is clear here: $+10.4$ points at $K{=}8$ ($p{=}2{\times}10^{-6}$) and $+14.7$ at $K{=}20$ ($p{=}10^{-8}$). \ours{} exceeds the embedding alone by $0.9$ points at $K{=}8$ ($p{=}3{\times}10^{-4}$), and audited properties would add a further $2.2$.

\subsection{Which personal parameters matter?}
\label{sec:adapters}

\begin{table}[t]
\centering
\small
\setlength{\tabcolsep}{2.8pt}
\begin{tabular}{lccc}
\toprule
 & \multicolumn{3}{c}{Brier $\downarrow$} \\
Personal adapter & $K{=}3$ & $8$ & $20$ \\
\midrule
Response calibration, no image & 0.548 & 0.521 & 0.503 \\
\midrule
Strictness & 0.517 & 0.493 & 0.480 \\
Property weights & 0.514 & 0.487 & 0.469 \\
Response calibration & 0.513 & 0.483 & 0.464 \\
\ours{} (weights + response) & \textbf{0.512} & \textbf{0.479} & \textbf{0.457} \\
\bottomrule
\end{tabular}

\caption{Personal parameterisations on the same people, answers and test images. The first row fits the response calibration without the scene; the others share the same group model of the scene.}
\label{tab:adapters}
\end{table}

\Cref{tab:adapters} compares ways of spending the $K$ answers. \emph{Strictness} adds one personal offset $b_p$ to \cref{eq:eta}. \emph{Property weights} fits $b_p$ and $\delta_p$. \emph{Response calibration} applies \cref{eq:response} to the group prediction and has no image-dependent personal parameter. \ours{} combines property weights with response calibration. Each has its priors tuned on training people.

\textbf{Personal calibration needs the scene.} Fit to the aid's answer distribution without the image, the response calibration scores $0.521$ at $K{=}8$, worse than \ours{} by $0.044$ (two-way CI $[0.025, 0.063]$, $85\%$ of people). On top of the scene model, a person's response style is the first thing the answers reveal: strictness alone recovers $0.057$ of the $0.070$ reduction at $K{=}8$, and response calibration, which also adapts the use of ``unsure'' and the confidence, adds $0.010$ at $K{=}8$ ($p{=}2{\times}10^{-4}$) and $0.016$ at $K{=}20$ ($p{=}7{\times}10^{-10}$).

\textbf{Property weights add to it.} Combining the two is better than response calibration alone by $0.004$ at $K{=}8$ ($p{=}0.001$) and $0.007$ at $K{=}20$ ($p{=}10^{-7}$, better for $70\%$ of people, in all $20$ partitions), better than property weights alone by $0.008$ ($p{=}0.03$) and $0.012$ ($p{=}6{\times}10^{-4}$), and makes more decisive statements than response calibration ($+1.1$ and $+1.6$ points, $p\le 10^{-5}$). The weights need more than three answers to show an effect. The decisive shares of all adapters are in the supplement.

\textbf{Are the properties the right personal axes?} Moving the personal weights to the first ten principal axes of the embedding, with the group model unchanged, is worse by $0.003$ at $K{=}8$ ($p{=}0.002$) and $0.005$ at $K{=}20$ ($p{=}10^{-5}$, $68\%$ of people). Moving them to ten random directions of the same $26$-dimensional space, which mix properties and embedding, matches \ours{} (difference at most $0.003$, $p\ge0.1$). The personal signal therefore lies in the joint span of the representation, which the leading embedding directions alone do not cover; the properties are one sufficient choice of axes, and the one whose parameters can be read, not the only one.

\subsection{Does it depend on the encoder?}
\label{sec:encoders}

\begin{table}[t]
\centering
\footnotesize
\setlength{\tabcolsep}{2.6pt}
\begin{tabular}{lccccc}
\toprule
 & \multicolumn{3}{c}{Brier, $K{=}20$} & \multicolumn{2}{c}{Decisive, $K{=}20$} \\
Encoder & Emb. & +Prop. & $\Delta$ & Emb. & +Prop. \\
\midrule
\multicolumn{6}{l}{\emph{Vision encoders (pooled output)}} \\
CLIP ViT-B/32 & 0.479 & 0.470 & $-0.009$ & 41.6 & 43.6 \\
CLIP ViT-B/16 & 0.493 & 0.478 & $-0.015$ & 39.5 & 42.6 \\
CLIP ViT-L/14 & 0.493 & 0.477 & $-0.015$ & 38.9 & 41.9 \\
DINOv2-B & 0.494 & 0.484 & $-0.010$ & 39.7 & 41.6 \\
DINOv2-L & 0.505 & 0.494 & $-0.011$ & 38.0 & 40.2 \\
SigLIP2-B & 0.469 & 0.462 & $-0.007$ & 43.5 & 45.7 \\
SigLIP2-So400m & 0.499 & 0.489 & $-0.010$ & 38.9 & 41.1 \\
ViT-B/16 (sup.) & 0.506 & 0.489 & $-0.018$ & 37.2 & 40.6 \\
\midrule
\multicolumn{6}{l}{\emph{Vision-language models (answer-position state)}} \\
LLaVA-1.5 7B & 0.479 & 0.470 & $-0.009$ & 41.9 & 43.9 \\
Qwen2.5-VL 7B & 0.480 & 0.471 & $-0.009$ & 43.1 & 44.4 \\
Cosmos-Reason1 7B & 0.476 & 0.466 & $-0.010$ & 42.4 & 44.7 \\
Qwen3.5 9B & 0.483 & 0.468 & $-0.015$ & 42.1 & 44.6 \\
Qwen3.6 27B & 0.470 & 0.464 & $-0.007$ & 43.9 & 45.1 \\
Qwen3.8 27B & 0.474 & 0.462 & $-0.012$ & 42.2 & 44.7 \\
Qwen3-VL 32B & 0.463 & 0.457 & $-0.007$ & 44.7 & 46.2 \\
Qwen3-VL 8B (ours) & 0.464 & 0.457 & $-0.006$ & 45.8 & 46.2 \\
\bottomrule
\end{tabular}

\caption{Sixteen frozen encoders as the scene embedding, without (Emb.) and with (+Prop.) the predicted properties, over $20$ partitions. Every $\Delta$ is significant ($p\le 2{\times}10^{-7}$, people paired). Decisive share in \% at $\alpha{=}0.05$.}
\label{tab:encoders}
\end{table}

We replace the scene embedding with each of sixteen frozen encoders, keeping the detector and the personal model fixed (\cref{tab:encoders}): eight vision encoders~\cite{radford2021learningtransferablevisualmodels,oquab2023dinov2,tschannen2025siglip2,dosovitskiy2021an} through their pooled output, and eight VLMs~\cite{liu2024improved,bai2025qwen25vl,azzolini2025cosmosreason1,bai2025qwen3vl} through the answer-position state of \cref{sec:embedding} (the supplement lists the checkpoints).

\textbf{Properties help] every encoder.} Adding the predicted properties improves all sixteen, by $0.006$ to $0.018$ Brier at $K{=}20$ and by $0.5$ to $3.4$ points of decisive share. The audit-supervised properties consistently add information that neither general-purpose vision embeddings nor the states of large VLMs expose to a model fit on fifty scenes.

\textbf{Which encoder.} Qwen3-VL gives the best scores, regardless of size with the $8$B model equaling the $32$B one. The other recent VLMs and the best vision encoder (SigLIP2-B) follow within $0.007$, and older VLMs are at the level of the vision encoders. We adopted Qwen3-VL-8B from this comparison. Choosing the encoder within each fold from the training-only tuning loss gives $0.486$ at $K{=}8$, with a Qwen3-VL model chosen in $68$ of $100$ folds; the supplement also repeats the analyses with a DINOv2 encoder fixed beforehand. Controls in the supplement show no improvement from adding the VLM's answer to a DINOv2 embedding, and worse performance when replacing the answer-position state with mean image-token features or a state extracted using a neutral prompt.

\subsection{What must perception get right?}
\label{sec:perception}

\begin{table}[t]
\centering
\small
\setlength{\tabcolsep}{4.5pt}
\begin{tabular}{lccc}
\toprule
Properties in the model & $K{=}8$ & $K{=}20$ & $p$ \\
\midrule
All predicted (\ours{}) & 0.479 & 0.457 &  \\
Audited tags & 0.480 & 0.457 & 0.14 \\
Audited severity & 0.467 & 0.445 & $6{\times}10^{-6}$ \\
Audited type & 0.474 & 0.449 & $4{\times}10^{-5}$ \\
Audited type and severity & 0.463 & 0.438 & $1{\times}10^{-10}$ \\
All audited & 0.463 & 0.438 & $2{\times}10^{-8}$ \\
\bottomrule
\end{tabular}

\caption{Oracle study (Brier): one block of predicted properties is replaced by its audited value. $p$: paired test against \ours{} at $K{=}20$.}
\label{tab:oracle}
\end{table}

\textbf{The gap is type and severity.} \Cref{tab:oracle} replaces one block of predicted properties at a time with the audited values. Audited severity gains $0.012$ at $K{=}20$ and audited type $0.009$, the two together account for the whole gap, and audited tags add nothing. On the survey images the detector identifies the type in $72\%$ of cases, with recall of $84\%$ for surface problems and $75\%$ for obstacles but $60\%$ for curb ramps ($3$ of $5$) and $33\%$ for missing curb ramps ($2$ of $6$), though both estimates rely on few images. Missing curb ramps also show the largest personal differences (\cref{sec:adapters}), suggesting a potentially important detection gap that needs confirmation on more examples. Tag AUCs range from $0.76$ to $0.96$. Predicted severity correlates with the audited value at $r{=}0.32$; two Project Sidewalk auditors who label the same object agree on its severity with within-type $r$ between $0.11$ and $0.44$ ($25{,}157$ pairs), so severity labels are themselves noisy. Linear probes on the VLM state recover these properties less well than the detector (type accuracy $56\%$), which is why the two sources are complementary. 

\textbf{Training data and geography.} Growing the detector's pool from $8.4$k to $14.9$k scenes raises type accuracy on held-out crops from $0.73$--$0.75$ to $0.79$--$0.80$ and lowers the personal Brier at $K{=}20$ from $0.460$ to $0.457$ ($p{=}7{\times}10^{-9}$); $18.1$k scenes add nothing.

\textbf{An unseen barrier type.} To test transfer beyond the scene distribution, we withhold one Project Sidewalk label type at a time: the group model is fit on the other three types and evaluated on the withheld one, with people split as before. Eight answers still lower the Brier score from $0.547$ to $0.504$ and twenty to $0.488$ ($p\le10^{-3}$), and the properties improve on the embedding alone by $0.006$ ($p{=}2{\times}10^{-10}$, $79\%$ of people). The supplement compares the adapters and the crowd oracle on this split.

\subsection{Who benefits}
\label{sec:who}

\textbf{Gains by aid} (supplement). With eight answers the gain is largest for walker and cane users ($0.115$ and $0.096$, $p\le 2{\times}10^{-5}$) and significant for manual wheelchair users ($p{=}0.002$), who are already below the crowd oracle. Motorized wheelchair and scooter users, whose answers depend more on the scene than on the person (supplement), need twenty answers to fall below the crowd oracle; the gain over $K{=}0$ is then significant for the former ($p{=}2{\times}10^{-4}$) but not for the $22$ scooter users ($p{=}0.07$, \cref{sec:limits}).

\textbf{People who disagree with the crowd gain more.} By distance of the eight calibration answers from others' average answer, the Brier gain is $0.039$ in the two quartiles closest to the crowd and $0.070$ and $0.157$ in the two farthest (supplement): those a population model serves worst gain most.

\subsection{A second population: rater strictness}
\label{sec:auditors}

The survey has $135$ people. To test whether few-shot personal calibration is specific to it, we use a second population in the same visual domain. Every Project Sidewalk label carries the identifier of the volunteer auditor who placed it and the severity that auditor gave, an individual and subjective judgment of a labelled object. The task mirrors \cref{sec:protocol}: given $K$ ratings from an auditor the model has not seen, predict the severity they give to other scenes. We use the same ordinal model with a personal strictness and five folds over auditors.

\textbf{With images.} For the $95$ auditors with at least $30$ ratings in our pool, the input is the embedding of the label crop and the label type. No panorama of a test auditor is used to fit the group model, and an auditor's calibration and test ratings come from different panoramas. Eight ratings improve the mean Brier by $\AuditorKEightGain{}$ (CI $[\AuditorKEightCILow{},\AuditorKEightCIHigh{}]$, $p{=}\AuditorKEightP{}$) and twenty by $\AuditorKTwentyGain{}$ ($p{=}8{\times}10^{-4}$; supplement). Here the personal signal is mostly one number, the auditor's strictness, and a prompt can carry it: Qwen3-VL-8B with the same eight ratings in context reaches $0.615$ after recalibration, level with the fitted model on the same items ($0.613$, $p{=}0.6$). On the mobility judgments, where the answer depends on the scene and the person together, it does not (\cref{sec:experiments}).

\textbf{Without images, at scale.} Personal strictness can be measured on far more auditors if the image is dropped. With label type and city as the only inputs, on $996$ auditors and $438$k ratings, one rating already helps ($-0.009$, $p{=}10^{-23}$), eight give $-0.040$ (better for $74\%$ of auditors) and twenty $-0.054$. Auditors are volunteers and not mobility-aid users, so both experiments support the stability of personal strictness, not claims about accessibility needs.

\section{Limitations}
\label{sec:limits}

\textbf{Scale and target.} The binding constraint is the $50$ scenes, the only public dataset of individual passability judgments by mobility-aid users that we know of. Repeated partitions cannot add visual diversity the pool does not have; the two-way bootstrap and the held-out barrier type measure this limit. Some properties, missing curb ramps among them, appear in as few as $6$ of the $50$ images. The $135$ people are adequate for the pooled comparisons; the smallest aid group ($22$ scooter users) is not, and its weaker significance reflects that. The encoder was adopted after comparing sixteen on this survey; nested selection costs $0.007$ (\cref{sec:encoders}). Judgments concern photographs, not traversals, and the detector is trained on audits from the survey cities.

\textbf{The guarantee is marginal.} The bound of \cref{sec:crc} holds on average over people and counts errors over all of a person's scenes. It does not control the error rate among the scenes that receive \pass{}, which ranges from $11\%$ for cane and walker users to $29\%$ for scooter users at $\alpha{=}0.05$ (supplement). Thresholds calibrated per aid restore the bound within each aid, but need at least $1/\alpha-1$ calibration people per group, more than the smaller groups have at $\alpha{=}0.05$.

\textbf{Use.} \ours{} is not a navigation safety system: the error among \pass{} statements is not bounded, and a deployment would need conservative, group-specific thresholds calibrated on more people than any public dataset has. Personal parameters reveal a person's aid and tolerance and should stay on the person's device. We use only the public release of Li et al., whose survey collected these judgments for research on personalised maps, and public Project Sidewalk labels, with auditor identifiers anonymised.

\section{Conclusion}
A sidewalk predictor should specify whose judgment it predicts. \ours{} adapts audit-supervised scene properties to a person from a few answers. On new people and scenes, eight answers cut the Brier score by $0.070$ and raise the decisive share at a $5\%$ risk bound from $27\%$ to $42\%$. These findings support separating shared scene understanding from personal interpretation to capture differences a single label misses. These predictions, combined with mobility-device-configuration~\cite{ahmed2026wheelchair}, could inform autonomous pickup and drop-off planning by identifying candidate locations to verify with the passenger. Human-centered accessibility perception must answer not only whether a sidewalk is passable, but for whom---and \ours{} shows how a few personal answers help bridge that gap.

{
    \small
    \bibliographystyle{ieeenat_fullname}
    \bibliography{refs}
}

\end{document}